\documentclass[conference]{IEEEtran}
\IEEEoverridecommandlockouts
\usepackage{cite}
\usepackage{amsmath,amssymb,amsfonts}
\usepackage{algorithmic}
\usepackage{graphicx}
\graphicspath{{./figures/}}
\usepackage{epstopdf}
\usepackage{textcomp}
\usepackage{xcolor}
\usepackage{booktabs}
\usepackage[nolist]{acronym}
\usepackage[colorinlistoftodos]{todonotes} 
\usepackage{booktabs,multirow} \usepackage[table]{xcolor}
\newcommand{\sd}[1]{{\color{gray}$\,\pm\,$#1}}
\definecolor{margshade}{gray}{0.92}
\newcommand{\nosh}{\cellcolor{white}}
\usepackage[caption=false,font=footnotesize]{subfig}

\usepackage{etoolbox}
\AtBeginEnvironment{table}{\footnotesize}
\AtBeginEnvironment{table*}{\footnotesize}

\title{Characterizing Behavioral Policy Differentiation under Aggregation and Credit Assignment in Co-Evolved Spiking Neural Network Ensembles}

\author{
\IEEEauthorblockN{Catherine Rodriquez}
\IEEEauthorblockA{
Louisiana State University\\
Baton Rouge, Louisiana, USA\\
crodr99@lsu.edu
}
\and
\IEEEauthorblockN{James Ghawaly Jr.}
\IEEEauthorblockA{
Louisiana State University\\
Baton Rouge, Louisiana, USA\\
jghawaly@lsu.edu
}
}

\begin{document}
\begin{acronym}

\acro{AI}{Artificial Intelligence}
\acro{SNN}{Spiking Neural Network}
\acro{CNN}{Convolutional Neural Network}
\acro{EONS}{Evolutionary Optimization for Neuromorphic Systems}
\acro{EO}{Evolutionary Optimization}
\acro{NEAT}{NeuroEvolution of Augmenting Topologies}

\acro{SNNEC}{Spiking Neural Network Ensemble Co-evolution}

\acro{RL}{Reinforcement Learning}
\acro{MARL}{Multi-Agent Reinforcement Learning}
\acro{MoE}{Mixture-of-Experts}
\acro{WTA}{Winner-Take-All}

\acro{DvD}{Diversity via Determinants}
\acro{EAV}{Expected Action Variation}
\acro{MMD}{Maximum Mean Discrepancy}
\acro{KED}{Key Expert Dependency}

\acro{RBF}{Radial Basis Function}
\acro{RFF}{Random Fourier Feature}

\end{acronym}
\maketitle

\begin{abstract}
Spiking neural network ensembles distribute intelligent capabilities across multiple members, enabling the efficiency of smaller models. Prior work has developed approaches to co-evolve these ensembles by optimizing marginal contribution fitness in an evolutionary optimization paradigm to encourage behavioral differentiation among ensemble members, with control tasks especially benefiting from co-evolution strategies. While it was speculated that co-evolution through marginal fitness encourages behavioral policy differentiation in control problems, this had not yet been tested. In this work, we study how credit assignment and ensemble aggregation methods influence member policy differentiation on a variety of standard control problems: CartPole, Acrobot, and MountainCar. For ensemble member credit assignment, we consider marginal and shared contribution fitness. For ensemble aggregation, we consider Winner-Take-All and Mean aggregation. We performed a $2\times2$ factorial experiment, generating metrics for policy differentiation, action disagreement, and state-action occupancy. 
We find that marginal contribution credit generally produced more behaviorally differentiated member policies than shared credit under both aggregation methods, with the exception of MountainCar under Mean aggregation. Unexpectedly, we also find that the aggregation technique itself has little effect on influencing differentiation, except on MountainCar where Winner-Take-All reduces it under shared credit but not under marginal credit.
Finally, qualitative analysis of individual evolved ensembles reveals intuitive divisions of control among members.

\end{abstract}


\section{Introduction}


\ac{EO} algorithms -- such as \ac{EONS}~\cite{schumaneons} -- offer a gradient-free approach to tuning both the topology and neuronal parameters of \acp{SNN}. This, combined with the ability to optimize for arbitrary fitness functions makes this approach attractive for neuromorphic computing applications. However, \ac{EO} exhibits super-exponential search scaling as the number of neurons in the \ac{SNN} increases. To combat this, prior work has demonstrated that ensembles of \acp{SNN} can be co-evolved by optimizing each member's marginal contribution to ensemble fitness, enabling better scaling performance on a variety of classification, regression, and control tasks~\cite{rodriquez2026coevolved}. Herein, we refer to this approach as \ac{SNNEC}. Notably, \ac{SNNEC} yielded especially significant improvements on control tasks; however, the effects it has on member-level policy differentiation were unstudied. We speculate that two separate factors within \ac{SNNEC} contribute to the emergence of member behavioral policy differentiation: \emph{member-level credit assignment} and \emph{ensemble decision aggregation}. 

Member-level credit assignment refers to the fitness credited to an ensemble's member. We consider two options, the first being marginal-contribution, which as its name suggests, measures the \emph{marginal} contribution of an individual member to an ensemble's overall performance as defined in prior work~\cite{rodriquez2026coevolved}. The second being \emph{shared} contribution, where each member receives the ensemble's overall fitness as its member-level fitness. 

Ensemble decision aggregation refers to the mechanism by which an ensemble aggregates member-level decisions into an overall decision. Here, we also consider two options: \emph{mean} aggregation and \emph{\ac{WTA}}. In mean aggregation, the average of the ensemble members' output is used as the ensemble's aggregate output. In \ac{WTA}, the most confident member's output is used as the ensemble's output.

In \ac{RL} literature, difference-reward methods have used counterfactual performance to isolate member contribution to a group's shared objective~\cite{castellini2022difference}. Likewise, work on hierarchical and expert-based \ac{RL} suggests that dividing control among multiple experts with a single winner's output can encourage the development of specialized policies \cite{hihnspecialization}. Together, these concepts motivate the expectation that marginal credit assignment and \ac{WTA} aggregation could encourage stronger differentiation among an ensemble's members in \ac{SNNEC}.

To understand how member-level credit assignment and ensemble aggregation impact member behavioral policy differentiation, we perform a $2\times2$ factorial experiment on three standard control problems: CartPole, Acrobot, and MountainCar. 
We also note that behavioral policy differentiation is likely a multidimensional property of ensemble organization. It may not always lead to higher task performance because behavioral policy and functional roles on a given task may not overlap. For example, while a particular member of an ensemble may visit different states than another, those states may not be important in the context of solving a particular task. For this reason, our experiments report a variety of literature metrics that each measure different qualities of behavioral policy differentiation to provide a more holistic interpretation.

\section{Related Work}

\subsection{Neuroevolution and Cooperative Co-Evolution}

Neuroevolution refers to \ac{EO} strategies that optimize neural network parameters, topology, or both. One of the most famous early approaches, \ac{NEAT}, performs joint evolution of network structure and connection weights for \ac{RL} control \cite{stanley2002neat}. For neuromorphic computing applications, \ac{EONS} extends \ac{EO} to \acp{SNN} intended for neuromorphic systems by directly optimizing them within the constraints of neuromorphic hardware\cite{schumaneons}.

Cooperative co-evolution extends \ac{EO} by decomposing a larger optimization problem into interacting subcomponents that evolve in separate populations \cite{potter2000cooperative}. This approach has been applied to ensembles of smaller \acp{SNN}, where marginal-contribution fitness evaluates each member according to the performance it uniquely adds to the ensemble~\cite{rodriquez2026coevolved}. Evolving multiple interacting networks also creates the possibility that members develop different behavioral roles and this is what we study herein.

\subsection{Credit Assignment and Individual Contribution}

Shared rewards measure collective performance but do not directly identify the contribution of individual components. In other words, they do not quantify \emph{member-level credit assignment}. Difference rewards offer a solution to this problem by comparing ensemble performance against a counterfactual in which an individual member's contribution is removed or replaced~\cite{castellini2022difference}. Related approaches such as counterfactual multi-agent (COMA) policy gradients use counterfactual baselines to improve individual credit assignment in \ac{MARL} \cite{foerster2018coma}.

Marginal-contribution fitness applies the same general principle at the ensemble-policy level. Each network is evaluated according to the performance difference caused by removing it from the ensemble. This changes the evolutionary signal from collective performance to individual contribution and provides a basis for examining whether credit assignment changes member behavioral policy differentiation.

\subsection{Behavioral Policy Differentiation and Expert Routing}

Specialized policies are common in architectures that divide a problem among multiple experts. \ac{MoE} models use gating mechanisms to assign inputs to different experts \cite{jacobs1991experts}, while sparse variants activate only a subset of experts for each input \cite{shazeer2017moe}. Similar ideas appear in \ac{RL} through hierarchical policies, options, and learned local experts \cite{sutton1999options,bacon2017optioncritic,hihnspecialization}.

Hard routing is particularly relevant to \ac{WTA}. Mo{\"E}T includes a routing variant in which a single expert produces each prediction \cite{vasic2022moet}. Other approaches allow roles to emerge rather than defining them in advance. ROMA and RODE learn roles for cooperative agents \cite{wang2020roma,wang2021rode}, while Poppy encourages complementary policies within an \ac{RL} population \cite{grinsztajn2023poppy}.

These approaches, in part, motivate the theoretical expectation that we examine in this work. Marginal credit assigns value to each member's contribution to the ensemble and \ac{WTA} gives individual members direct control over their decisions. Taken together, these mechanisms create a setting in which distinct member roles may be more likely to emerge. Therefore, we examine how each combination shapes the resulting behavioral policy differentiation profile instead of assuming that marginal credit and \ac{WTA} must always produce greater differentiation.

\subsection{Behavioral Diversity and Functional Specialization}

Behavioral diversity is related to specialization but does not necessarily establish distinct functional roles. Methods such as DIAYN show that distinguishable behaviors can be learned directly \cite{eysenbach2019diayn}, while diversity measures capture different properties of policy populations \cite{liu2022unified}.


Regarding metrics for evaluating diversity, \ac{DvD} captures population-level differentiation and redundancy \cite{dvdmetric}, while \ac{EAV} measures action variation under common states \cite{eavmetric}. Occupancy-based measures compare the state-action distributions induced over trajectories, with \ac{MMD} providing one approach for quantifying these differences \cite{hedayatian2026autoqd}. The various perspectives of these metrics on measuring policy diversity motivate our treatment of behavioral policy differentiation as a profile across several measures rather than a single score.

\section{Methodology}




\subsection{SNNEC}

\ac{SNNEC} represents the control policy as an ensemble of three \acp{SNN} evolved in separate \ac{EONS} populations and is described in detail in prior work~\cite{rodriquez2026coevolved}. Genetic operations are performed within each population, while fitness evaluation couples the populations through ensemble performance. Candidate networks are grouped by index, such that candidate $j$ from each population forms a three-member ensemble.

At each environment step, the current observation is encoded as input events and evaluated by each ensemble member. A fresh finite-duration \ac{SNN} simulation is performed for every decision so that neural state is not carried between environment steps. The resulting output activity is decoded into action-level information and combined according to the selected aggregation strategy.

\subsection{Credit Assignment}




We consider two approaches to member-level credit assignment: shared and marginal-contribution fitness.

Under shared credit, each member receives the ensemble's overall fitness. For a given ensemble, the mean episodic return is assigned to all three members, meaning that each network is rewarded according to how well the ensemble performs as a whole.

Marginal-contribution credit instead rewards each member based on how much it contributes to the ensemble's performance. For ensemble $G$ and member $i$,

\begin{equation}
\Delta_i(G)=J(G)-J(G\setminus\{i\}),
\end{equation}


where $J(G)$ is the mean episodic return of the full ensemble and $J(G\setminus\{i\})$ is the mean return after removing member $i$. The full and reduced ensembles are evaluated using the same episode seeds and aggregation strategy. Negative contributions are retained when removing a member improves ensemble performance.

This provides each member with a fitness value based on its individual contribution rather than assigning the same ensemble-level fitness to all members.

\subsection{Aggregation Strategies}




We consider two approaches to ensemble decision aggregation: Mean and \ac{WTA}.

Under Mean aggregation, the outputs of all three members contribute to the ensemble decision. For spike-count decoding, each member produces an output spike count for every available action. These outputs are averaged across the ensemble, and the action with the largest mean count is selected. Member outputs are not normalized before averaging, allowing differences in firing activity to affect the ensemble decision.

Under \ac{WTA} aggregation, a single member determines the ensemble action. Each member's output counts are first normalized across the available actions. Each member then proposes its preferred action and is assigned a confidence value based on the difference between its two largest normalized outputs,

\begin{equation}
q_i(s)=p_{i,(1)}(s)-p_{i,(2)}(s).
\end{equation}


The member with the highest confidence ``wins,'' and its preferred action is used as the ensemble action. In this way, Mean combines the outputs of all members, while \ac{WTA} gives control of each decision to a single member. Environment-specific decoding and confidence calculations are described in the experimental setup.

\subsection{Factorial Design}


We vary member-level credit assignment and ensemble decision aggregation independently in a $2\times2$ factorial design. This produces four conditions: Mean + Shared, Mean + Marginal, \ac{WTA} + Shared, and \ac{WTA} + Marginal.

This allows us to examine the effect of changing how individual members are rewarded separately from changing how their outputs are used to form the ensemble decision, as well as how the two mechanisms interact.

\subsection{Behavioral Policy Differentiation Metrics}


We use three metrics to measure different forms of member behavioral policy differentiation. Rather than combining them into a single score, we consider each separately since they capture different aspects of how the ensemble members behave.

\subsubsection{Diversity via Determinants}

\ac{DvD} measures how behaviorally distinct the member policies are as a group \cite{dvdmetric}. Each member is evaluated independently on the same set of reference states, and its deterministic actions are used to form a behavioral representation. A \ac{RBF} similarity matrix is then constructed from these representations, with diversity given by

\begin{equation}
\mathrm{DvD}=\det(K).
\end{equation}


Values near zero indicate that two or more members behave similarly, while larger values indicate a more distinct set of member policies. A value of exactly zero indicates identical behavior between at least two members. We use the same kernel bandwidth across conditions within each environment.

\subsubsection{Expected Action Variation}

\ac{EAV} measures how often ensemble members choose different actions when given the same state \cite{eavmetric}. Since the member policies are deterministic, \ac{EAV} is calculated as the average pairwise disagreement rate across the same reference states used for \ac{DvD}. Higher values indicate greater action-level disagreement among the members.

\subsubsection{State-Action Occupancy MMD}

Occupancy \ac{MMD} measures whether members move through the environment differently over time. Each member independently controls the environment using the same evaluation seed set, and its resulting state-action transitions are collected. States are standardized using pooled statistics within each environment, while actions are represented using one-hot vectors.

Differences between member trajectories are then measured using an \ac{RBF}-kernel approximation to squared \ac{MMD} with \acp{RFF} \cite{hedayatian2026autoqd}. Each member's state-action transitions are mapped into a shared feature space and averaged to produce an occupancy representation. We report the mean pairwise squared distance between the three member representations. Higher values indicate that members produce more differentiated state-action trajectories.






\begin{table*}[h!]
\centering
\caption{Environment and training settings used for the behavioral policy differentiation experiments.}
\label{tab:environment-settings}

\renewcommand{\arraystretch}{1.2}

\begin{tabular*}{\textwidth}{@{\extracolsep{\fill}}lcccccccc@{}}
\toprule
Environment & Obs. & Actions & SNN Inputs & Decoder & Runtime & Pop. Size & Generations & Train Episodes \\
\midrule
CartPole-v1    & 4 & 2 & 4  & Spike count & 100 & 60  & 100 & 5  \\
Acrobot-v1     & 6 & 3 & 6  & Spike count & 100 & 60  & 100 & 5  \\
MountainCar-v0 & 2 & 3 & 14 & First spike & 150 & 120 & 200 & 10 \\
\bottomrule
\end{tabular*}
\end{table*}

\subsection{Statistical Analysis}

For each behavioral policy differentiation metric, we examine how credit assignment and aggregation affect the resulting ensemble behavior within each environment. We calculate the effect of marginal credit under both Mean and \ac{WTA} aggregation, the effect of \ac{WTA} under both Shared and Marginal credit, and the interaction between the two factors. All 50 runs of each condition are included.

Because the same run IDs are used across all four conditions within an environment, each contrast is computed from the 50 within-run differences between two conditions, and its test statistic is the mean of those differences. We test each contrast with a two-sided paired permutation test, randomly flipping the sign of each within-run difference over 100{,}000 permutations. The four contrasts for each metric and environment yield 36 tests. For the interaction, we compute each run's difference of differences, $(\text{marginal} - \text{shared})_{\text{WTA}} - (\text{marginal} - \text{shared})_{\text{Mean}}$, and test its mean with the same procedure, yielding nine tests. Pointwise 95\% intervals for the interactions are obtained from 100{,}000 run-ID block-bootstrap resamples. Benjamini--Hochberg correction is applied separately to the 36 contrasts and the nine interactions. Each environment is analyzed separately and metric values are not pooled across tasks.


\section{Experimental Setup}

\subsection{Control Environments}

We evaluate the four experimental conditions on CartPole-v1, Acrobot-v1, and MountainCar-v0 from the Gym/Gymnasium benchmark family \cite{brockman2016openaigym,towers2026gymnasium}. These environments span pole stabilization, underactuated swing-up, and momentum-based control \cite{barto1983cartpole,spong1995acrobot,sutton1996generalization}. While these environments are relatively simple, all are nonlinear dynamical systems that more easily enable interpretation than more complex, higher-dimensional environments. For rewards, we use those provided by the environment without additional shaping.


Table~\ref{tab:environment-settings} lists the training settings used for each environment. The reported runtime is the number of \ac{SNN} simulation-time units used for each network evaluation.

\subsection{Observation Encoding and Action Decoding}

For all environments, nonfinite observation values are replaced with zero, clipped to fixed environment-specific bounds, and normalized to $[0,1]$. Positive normalized values generate amplitude-coded input events at simulation time 1 using an environment-specific gain.

CartPole and Acrobot use their physical observation vectors directly with an input gain of 3. MountainCar uses gain 4 and augments its two physical observations with the four preceding actions, represented as three-dimensional one-hot vectors, for 14 total inputs.

CartPole and Acrobot use spike-count decoding. Under Mean aggregation, member output vectors are averaged before action selection. Under \ac{WTA}, each member proposes an action and is assigned confidence from the difference between its two largest normalized output scores.

MountainCar uses first-spike decoding. Under Mean aggregation, the decoded member actions are combined by vote, while \ac{WTA} confidence is based on first-spike latency,

\[
q_i=\max\left(0,1-\frac{t_i}{T}\right),
\]

where $t_i$ is the first output-spike time and $T$ is the \ac{SNN} runtime.

\subsection{Evolutionary Training}



Each environment uses three separately evolving \ac{EONS} populations, with one network from each population forming a three-member ensemble. Networks begin with 12 hidden nodes and 32 initial edge-addition attempts, with a maximum of 160 hidden nodes allowed during evolution.

Within each population, three elite networks are retained between generations. Mutation is applied with probability 0.9, crossover with probability 0.5, and four mutation operations are attempted during each mutation step. Add-node, add-edge, delete-node, and delete-edge operations are sampled with relative weights of 0.5, 0.75, 0.25, and 0.25, respectively.

The evolved networks use nonleaky integrate-and-fire neurons. Neuron thresholds, inhibitory polarity, connection weights, and delays are all allowed to evolve.

\subsection{Evaluation and Policy Selection}

For each evolutionary run, the highest-scoring ensemble from the final training generation is selected for evaluation. Each selected ensemble contains one network from each of the three evolving populations.


The selected ensemble is then evaluated over 100 fixed test episodes. The same episode seeds are used across all four experimental conditions within an environment. These same selected networks are used for the behavioral policy differentiation metrics reported in this work.

\begin{table*}[t!]
\centering
\caption{Specialization metrics across all 50 runs per condition. Values are mean \sd{sample standard deviation}. Arrows indicate the direction of greater behavioral differentiation. Shaded rows use marginal-contribution credit. Bold marks the most differentiated specialization value within each environment and does not by itself indicate statistical significance; see Tables~\ref{tab:simple-effects} and~\ref{tab:interactions}. Median and third-quartile returns are calculated across the 50 run-level mean test returns.}
\label{tab:metrics}
\setlength{\tabcolsep}{4.5pt}
\begin{tabular}{lll rrr rr}
\toprule
Environment & Aggregation & Credit & DvD $\uparrow$ & EAV $\uparrow$ & Occ.\ MMD $\uparrow$ & Median Return & Q3 Return \\
\midrule
 & & Shared & 0.056\sd{0.118} & 0.416\sd{0.174} & 0.313\sd{0.141} & 498.04 & 500.00 \\
\rowcolor{margshade}
\nosh & \nosh\multirow{-2}{*}{Mean} & Marginal & \textbf{0.174}\sd{0.173} & \textbf{0.518}\sd{0.174} & \textbf{0.425}\sd{0.170} & 496.93 & 500.00 \\
\cmidrule(lr){2-8}
 & & Shared & 0.083\sd{0.120} & 0.383\sd{0.175} & 0.308\sd{0.143} & 497.46 & 500.00 \\
\rowcolor{margshade}
\nosh\multirow{-4}{*}{CartPole} & \nosh\multirow{-2}{*}{WTA} & Marginal & 0.096\sd{0.139} & 0.491\sd{0.164} & 0.374\sd{0.170} & 495.04 & 500.00 \\
\midrule

 & & Shared & 0.095\sd{0.145} & 0.555\sd{0.191} & 0.449\sd{0.120} & -89.83 & -83.02 \\
\rowcolor{margshade}
\nosh & \nosh\multirow{-2}{*}{Mean} & Marginal & \textbf{0.177}\sd{0.156} & 0.641\sd{0.140} & \textbf{0.505}\sd{0.108} & -106.53 & -90.88 \\
\cmidrule(lr){2-8}
 & & Shared & 0.096\sd{0.159} & 0.547\sd{0.211} & 0.443\sd{0.097} & -88.64 & -83.46 \\
\rowcolor{margshade}
\nosh\multirow{-4}{*}{Acrobot} & \nosh\multirow{-2}{*}{WTA} & Marginal & 0.165\sd{0.173} & \textbf{0.649}\sd{0.164} & 0.486\sd{0.121} & -93.00 & -86.02 \\
\midrule

 & & Shared & \textbf{0.240}\sd{0.046} & \textbf{0.666}\sd{0.058} & \textbf{0.374}\sd{0.057} & -106.11 & -105.79 \\
\rowcolor{margshade}
\nosh & \nosh\multirow{-2}{*}{Mean} & Marginal & 0.209\sd{0.075} & 0.658\sd{0.068} & 0.354\sd{0.083} & -107.88 & -106.18 \\
\cmidrule(lr){2-8}
 & & Shared & 0.088\sd{0.100} & 0.578\sd{0.085} & 0.334\sd{0.109} & -109.83 & -109.64 \\
\rowcolor{margshade}
\nosh\multirow{-4}{*}{MountainCar} & \nosh\multirow{-2}{*}{WTA} & Marginal & 0.176\sd{0.134} & 0.614\sd{0.154} & 0.362\sd{0.106} & -109.82 & -109.65 \\
\bottomrule
\end{tabular}
\end{table*}

\subsection{Run Cohorts}

The primary analysis uses fixed run IDs run000--run049 for each condition and environment. The same run IDs are used across all four experimental conditions, preserving the matched seed structure and allowing conditions to be compared directly.

\subsection{Behavioral Policy Differentiation Analysis Procedure}



The three behavioral policy differentiation metrics are computed from the three members in each selected ensemble after training.

For \ac{DvD} and \ac{EAV}, five fixed-seed reference episodes are generated for every selected ensemble within an environment. The resulting observations are pooled across the cohort, and 1,000 probe states are sampled from this shared set. All three members of every ensemble are then evaluated on the same 1,000 probes. The same probes are used for both \ac{DvD} and \ac{EAV}.

For occupancy \ac{MMD}, each member is evaluated independently for 100 episodes using the same episode seeds. Physical observations are standardized using the pooled trajectories within each environment, and member trajectories are represented using 1,024 shared \acp{RFF}.


Any shared probe sets, normalization values, kernel parameters, and feature mappings are determined separately for each environment.

\section{Results and Discussion}

Table~\ref{tab:metrics} summarizes the behavioral policy differentiation metrics across CartPole, Acrobot, and MountainCar for all 50 runs of each condition. Since each metric captures a different aspect of behavioral differences, we consider them together rather than using any single metric to rank the four conditions. Table~\ref{tab:simple-effects} reports the effect of each factor with the other held fixed, and Table~\ref{tab:interactions} reports the aggregation-by-credit interactions.

\begin{table*}[t!]
\centering
\caption{Simple-effect comparisons across all 50 runs per condition. Each contrast compares two conditions differing in one factor, with the other factor held fixed (shown in parentheses). $\Delta$ is the mean of the 50 within-run differences (first condition minus second). Tests are two-sided paired sign-flip permutation tests (100{,}000 permutations), and $p_{\mathrm{BH}}$ is Benjamini--Hochberg adjusted across all 36 comparisons. Bold indicates significance at a 5\% false discovery rate.}
\label{tab:simple-effects}
\setlength{\tabcolsep}{4pt}
\begin{tabular}{ll rr rr rr}
\toprule
 & & \multicolumn{2}{c}{DvD} & \multicolumn{2}{c}{EAV} & \multicolumn{2}{c}{Occ.\ MMD} \\
\cmidrule(lr){3-4} \cmidrule(lr){5-6} \cmidrule(lr){7-8}
Environment & Contrast & $\Delta$ & $p_{\mathrm{BH}}$ & $\Delta$ & $p_{\mathrm{BH}}$ & $\Delta$ & $p_{\mathrm{BH}}$ \\
\midrule
\multirow{4}{*}{CartPole}
 & Marginal $-$ Shared (Mean) & $\mathbf{+0.1185}$ & \textbf{0.0055} & $\mathbf{+0.1028}$ & \textbf{0.0388} & $\mathbf{+0.1119}$ & \textbf{0.0157} \\
 & Marginal $-$ Shared (WTA)  & $+0.0134$ & 0.7679 & $\mathbf{+0.1079}$ & \textbf{0.0157} & $+0.0656$ & 0.1168 \\
 & WTA $-$ Mean (Shared)      & $+0.0271$ & 0.3917 & $-0.0326$ & 0.5036 & $-0.0047$ & 0.8979 \\
 & WTA $-$ Mean (Marginal)    & $-0.0780$ & 0.0774 & $-0.0276$ & 0.5878 & $-0.0509$ & 0.2871 \\
\midrule
\multirow{4}{*}{Acrobot}
 & Marginal $-$ Shared (Mean) & $\mathbf{+0.0823}$ & \textbf{0.0388} & $+0.0855$ & 0.0554 & $+0.0561$ & 0.0756 \\
 & Marginal $-$ Shared (WTA)  & $+0.0697$ & 0.0901 & $\mathbf{+0.1023}$ & \textbf{0.0403} & $+0.0436$ & 0.1540 \\
 & WTA $-$ Mean (Shared)      & $+0.0007$ & 0.9832 & $-0.0084$ & 0.8667 & $-0.0065$ & 0.8489 \\
 & WTA $-$ Mean (Marginal)    & $-0.0120$ & 0.8164 & $+0.0084$ & 0.8489 & $-0.0191$ & 0.5216 \\
\midrule
\multirow{4}{*}{MountainCar}
 & Marginal $-$ Shared (Mean) & $\mathbf{-0.0310}$ & \textbf{0.0455} & $-0.0084$ & 0.6758 & $-0.0199$ & 0.2528 \\
 & Marginal $-$ Shared (WTA)  & $\mathbf{+0.0881}$ & \textbf{0.0025} & $+0.0365$ & 0.2240 & $+0.0278$ & 0.3032 \\
 & WTA $-$ Mean (Shared)      & $\mathbf{-0.1521}$ & \textbf{0.0002} & $\mathbf{-0.0886}$ & \textbf{0.0002} & $-0.0407$ & 0.0763 \\
 & WTA $-$ Mean (Marginal)    & $-0.0330$ & 0.2240 & $-0.0437$ & 0.1168 & $+0.0070$ & 0.8164 \\
\bottomrule
\end{tabular}
\end{table*}

\begin{table}[t!]
\centering
\caption{Aggregation $\times$ credit interactions across all 50 runs per condition. The interaction is the marginal-credit effect under WTA minus the marginal-credit effect under Mean, computed within each run and averaged. Intervals are pointwise 95\% run-ID block-bootstrap percentile intervals (100{,}000 draws) and are not adjusted for multiplicity. $p_{\mathrm{BH}}$ is from a two-sided paired sign-flip permutation test, Benjamini--Hochberg adjusted across the nine interaction tests. Bold indicates significance at a 5\% false discovery rate. MMD denotes occupancy MMD.}
\label{tab:interactions}
\setlength{\tabcolsep}{3pt}
\begin{tabular}{llrcr}
\toprule
Environment & Metric & Interaction & 95\% CI & $p_{\mathrm{BH}}$ \\
\midrule
\multirow{3}{*}{CartPole}
 & DvD & $-0.1051$ & $[-0.1884, -0.0229]$ & 0.0783 \\
 & EAV & $+0.0050$ & $[-0.0916, +0.1039]$ & 0.9207 \\
 & MMD & $-0.0463$ & $[-0.1416, +0.0505]$ & 0.6292 \\
\midrule
\multirow{3}{*}{Acrobot}
 & DvD & $-0.0127$ & $[-0.0963, +0.0705]$ & 0.8643 \\
 & EAV & $+0.0168$ & $[-0.0778, +0.1105]$ & 0.8643 \\
 & MMD & $-0.0125$ & $[-0.0829, +0.0551]$ & 0.8643 \\
\midrule
\multirow{3}{*}{MountainCar}
 & DvD & $\mathbf{+0.1191}$ & $[+0.0711, +0.1681]$ & \textbf{0.0002} \\
 & EAV & $+0.0449$ & $[-0.0027, +0.0917]$ & 0.1584 \\
 & MMD & $+0.0477$ & $[+0.0016, +0.0931]$ & 0.1443 \\
\bottomrule
\end{tabular}
\end{table}

\subsection{Effect of Member-level Credit Assignment}
\label{sec:credit}

Marginal-contribution credit generally produced more behaviorally differentiated members than shared credit under both aggregation methods (Table~\ref{tab:simple-effects}). Under Mean aggregation (the original approach in SNNEC), marginal credit increased all three measures in CartPole: DvD by 0.119 ($p_{\mathrm{BH}}=0.006$), EAV by 0.103 ($p_{\mathrm{BH}}=0.039$), and occupancy MMD by 0.112 ($p_{\mathrm{BH}}=0.016$). In Acrobot, it increased DvD by 0.082 ($p_{\mathrm{BH}}=0.039$), with increases in EAV and occupancy MMD that did not reach significance. Under WTA aggregation, marginal credit increased EAV in CartPole by 0.108 ($p_{\mathrm{BH}}=0.016$) and in Acrobot by 0.102 ($p_{\mathrm{BH}}=0.040$), and DvD in MountainCar by 0.088 ($p_{\mathrm{BH}}=0.003$). Across both aggregation methods, the estimated effect of marginal credit was positive in all 12 comparisons in CartPole and Acrobot.

MountainCar under Mean aggregation was the exception, where marginal credit slightly reduced DvD by 0.031 ($p_{\mathrm{BH}}=0.046$), with small, nonsignificant reductions in EAV and occupancy MMD. One possible explanation is that MountainCar's Mean aggregation operates as a majority vote over decoded actions. Under a three-member vote, two members that agree determine the ensemble action whenever the third disagrees, so removing either one can change the outcome. Therefore, marginal credit can reward agreement between members under voting, whereas under spike-count averaging, a member that duplicates its partners has less influence on the ensemble's decisions and as a result, receives less credit.

Across environments, the differences appeared mainly in DvD and EAV, which compare member actions on states visited by the \emph{intact} ensemble. Occupancy MMD, however, showed fewer significant differences. MMD is computed from each member acting alone, and members are evolved to contribute to an \emph{ensemble} rather than to solve the task individually. As a result, MMD largely reflects how members behave outside of the conditions they were trained for, including how they fail. We therefore treat DvD and EAV as the primary evidence of differentiation within the ensemble, and occupancy MMD as a secondary measure of standalone behavior.

\subsection{Aggregation and Its Interaction with Credit Assignment}
\label{sec:aggregation}
On its own, aggregation had minimal effect on member differentiation, as seen in Table~\ref{tab:simple-effects}. The only exception was MountainCar under shared credit, where WTA reduced DvD by 0.152 and EAV by 0.089 relative to Mean aggregation ($p_{\mathrm{BH}}<0.001$ for both). This reduction disappeared when using marginal credit assignment, where WTA and Mean aggregation produced similar values on all three measures. The corresponding interaction was significant for DvD ($+0.119$, $p_{\mathrm{BH}}<0.001$; Table~\ref{tab:interactions}), but no other interaction was significant after correction.

Contrary to our expectations based on the literature~\cite{jacobs1991experts}, routing each decision through a single member did not by itself increase differentiation. One potential reason is that the competitive selection studied in prior work directly determines which expert receives credit, such that only the expert selected for a given input is updated~\cite{jacobs1991experts, grinsztajn2023poppy}. In our design, WTA determines which member performs an action, but fitness is still computed from whole-episode returns under both credit schemes. A member is therefore not credited specifically for the decisions it controls. Under WTA with shared credit, every member receives the ensemble's return regardless of whether it is selected. As such, members that rarely act or that duplicate the selected member do not face pressure to differentiate. Marginal credit restores that pressure by giving lower fitness to members that do not change the ensemble's decisions, which is consistent with the recovery of differentiation under WTA in MountainCar. Because MountainCar also uses first-spike decoding and differs in both aggregation implementations, the present results cannot separate the effect of routing from these environment-specific factors. Overall, credit assignment appears to be the primary influence on member differentiation, with aggregation having little effect except in combination with shared credit in MountainCar.

\subsection{Qualitative Analysis}

\begin{figure}[t]
\centering
\subfloat[Member selected at each observed state.\label{fig:acrobot-control-map}]{%
  \includegraphics[width=\columnwidth]{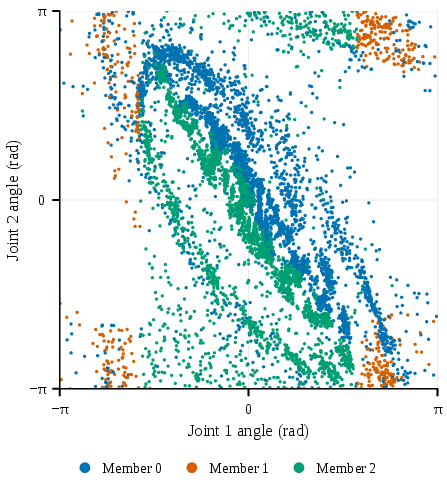}}
\\
\subfloat[Member selected at each step of one episode.\label{fig:acrobot-control-time}]{%
  \includegraphics[width=\columnwidth]{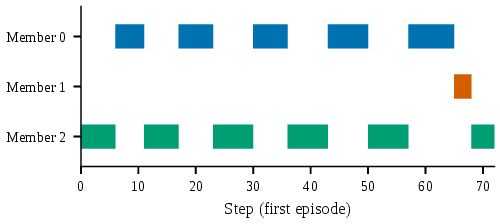}}
\caption{Member selection in a WTA + Marginal Acrobot ensemble.}
\label{fig:acrobot-member-control}
\end{figure}

\begin{figure}[t]
\centering
\subfloat[Member selected at each observed state.\label{fig:mountaincar-control-map}]{%
  \includegraphics[width=\columnwidth]{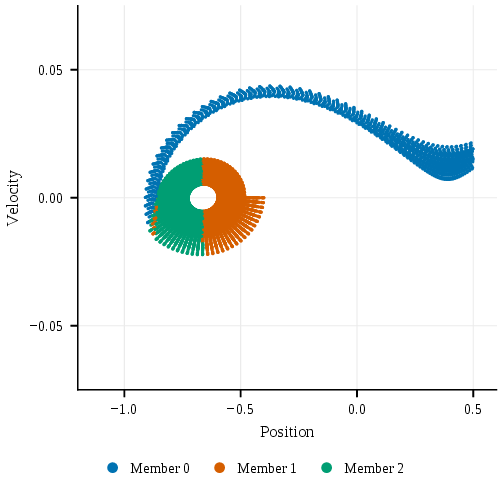}}
\\
\subfloat[Member selected at each step of one episode.\label{fig:mountaincar-control-time}]{%
  \includegraphics[width=\columnwidth]{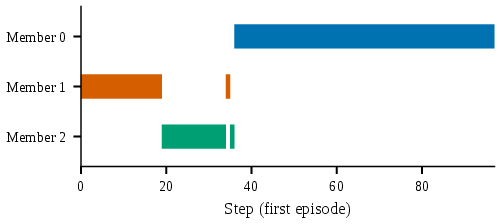}}
\caption{Member selection in a WTA + Marginal MountainCar ensemble.}
\label{fig:mountaincar-member-control}
\end{figure}

Figure~\ref{fig:acrobot-member-control} visualizes policy differentiation among members for a WTA + Marginal ensemble on Acrobot. Computed over 100 episodes, members 0 and 2 contribute 46.7\% and 47.5\% of decisions, respectively. As Figure~\ref{fig:acrobot-control-time} shows, members 0 and 2 alternate control in regular blocks of several steps throughout most of the episode and occupy parallel but offset regions of joint-angle space (Fig.~\ref{fig:acrobot-control-map}). Because swinging the Acrobot up requires alternating the direction of applied torque, this pattern is consistent with the two members handling different phases of the swing. Member 1 is selected far less often, supplying 5.8\% of actions, and only in a distinct region of large joint angles that the ensemble reaches late in the episode, once the swing has built up enough energy. 

Similarly, Figure~\ref{fig:mountaincar-member-control} shows a WTA + Marginal MountainCar ensemble, where we can see that control is divided into sequential phases. Members 1 and 2 control the initial oscillations, with member 1 selected on the right side of each oscillation and member 2 on the left side (Fig.~\ref{fig:mountaincar-control-map}). Once the car has gained enough momentum, member 0 takes over and controls the final swing until the car reaches the goal, remaining selected for the rest of the episode (Fig.~\ref{fig:mountaincar-control-time}). Members 0, 1, and 2 supply 13.4\%, 41.4\%, and 45.2\% of actions, respectively, computed on 100 episodes. In contrast to the alternating control in the Acrobot example, this ensemble divides the task over time, with two members building momentum and a third using it to reach the goal.

\subsection{Limitations}

This study has several limitations that should be understood when interpreting the results. First, the metrics used here capture behavioral differentiation, not functional roles. DvD and EAV show that members select different actions on the same states, but do not necessarily indicate whether those differences are complementary or necessary for the ensemble's success. Additionally, occupancy MMD is computed from members acting alone, a setting they were not evolved for. For this reason, it largely reflects standalone behavior rather than behavior within the ensemble. As previously stated, we view MMD as a secondary metric to DvD and EAV. 

We also note that the evaluation is limited to three low-dimensional control tasks with three-member ensembles. Co-evolving ensembles across multiple conditions with enough runs for reliable statistics requires substantial computation. The three control tasks we selected are nonlinear dynamical systems that require nontrivial control strategies, such as stabilizing an unstable equilibrium in CartPole, pumping energy into an underactuated system in Acrobot, and building momentum in MountainCar. Additionally, the tasks were chosen in part because their low-dimensional state spaces allow member behavior to be inspected directly, as in the examples shown in Figures~\ref{fig:acrobot-member-control} and~\ref{fig:mountaincar-member-control}. Extending the analysis to larger ensembles and higher-dimensional tasks remains an interesting direction for future work. 

\section{Conclusion}

This work examined how credit assignment and aggregation affect behavioral differentiation among members of co-evolved SNN ensembles, using a $2\times2$ design that crossed shared and marginal-contribution credit with Mean and WTA aggregation across CartPole, Acrobot, and MountainCar. Across all evolutionary runs, marginal-contribution credit generally produced members that selected different actions more often than shared credit, under both aggregation methods, with the exception of MountainCar under Mean aggregation. Aggregation alone had little effect on differentiation, with the exception of MountainCar, where WTA with shared credit produced less differentiated members, a reduction that did not occur under marginal credit.

These results suggest that credit assignment is the primary influence on whether ensemble members differentiate, while aggregation has little effect on its own. Routing each decision through a single member did not by itself produce more differentiated members, consistent with prior work in which competitive selection determines which expert receives credit rather than only which expert acts. At the same time, individual ensembles showed clear divisions of control under WTA: in Acrobot, we observe that two members alternated control throughout the swing while a third acted only in a late, high-energy phase, and in MountainCar, members took over in sequence, with two building momentum and a third carrying the car to the goal.

Several directions follow from these findings. Assigning credit to the decisions each member controls, rather than to whole-episode returns, may strengthen differentiation under WTA aggregation. Quantifying how control is divided across all runs, comparing against independently evolved networks, and extending the analysis to larger ensembles and more complex tasks would clarify how general these patterns are and how they arise.

\bibliographystyle{IEEEtran}
\bibliography{references}

\end{document}